\documentclass[pdflatex,sn-mathphys]{sn-jnl}

\jyear{2021}%

\theoremstyle{thmstyleone}%
\theoremstyle{thmstyletwo}%

\theoremstyle{thmstylethree}%

\usepackage{adjustbox}
\usepackage{multirow}
\usepackage{bbold}
\usepackage{siunitx}

\usepackage{colortbl}
\usepackage{hhline}
\usepackage{float}
\usepackage{diagbox}

\usepackage[normalem]{ulem}
\usepackage{color}
\usepackage{booktabs}
\usepackage{bbm}
\usepackage{makecell}
\usepackage{rotating}
\usepackage{hyperref}
\usepackage{amsmath}

\begin{document}

\title[PDZSeg: Dissection Zone Segmentation]{PDZSeg: Adapting the Foundation Model for Dissection Zone Segmentation with Visual Prompts in Robot-assisted Endoscopic Submucosal Dissection}


\author[1,2]{\fnm{Mengya} \sur{Xu}}
\equalcont{These authors contributed equally to this work.}
\author[3]{\fnm{Wenjin} \sur{Mo}}
\equalcont{These authors contributed equally to this work.}
\author[1,2]{\fnm{Guankun} \sur{Wang}}
\author[1,2]{\fnm{Huxin} \sur{Gao}}
\author[1,2]{\fnm{An} \sur{Wang}}
\author[4]{\fnm{Zhen} \sur{Li}}
\author[4]{\fnm{Xiaoxiao} \sur{Yang}}
\author*[1,2]{\fnm{Hongliang} \sur{Ren}}\email{hlren@ieee.org}


\affil*[1]{\orgdiv{Dept. of Electronic Engineering}, \orgname{CUHK}, \orgaddress{\city{Hong Kong}, \country{China}}}

\affil[2]{\orgdiv{Shenzhen Research Institute}, \orgname{CUHK}, \orgaddress{\city{Shenzhen}, \country{China}}}

\affil[3]{\orgdiv{Dept. of Computer Science and Engineering}, \orgname{Sun Yat-sen University}, \orgaddress{\city{Guangzhou}, \country{China}}}

\affil[4]{\orgdiv{Dept. of Gastroenterology}, \orgname{Qilu Hospital of Shandong University}, \orgaddress{\city{Jinan}, \country{China}}}




\abstract{\textbf{Purpose:} 
The intricate nature of endoscopic surgical environments poses significant challenges for the task of dissection zone segmentation. Specifically, the boundaries between different tissue types lack clarity which can result in significant segmentation errors, as the models may misidentify or overlook object edges altogether. Thus, the goal of this work is to achieve the precise dissection zone suggestion under these challenges during endoscopic submucosal dissection (ESD) procedures and enhance the overall safety of ESD. 

\textbf{Methods:} 
We introduce a Prompted-based Dissection Zone Segmentation (PDZSeg) model, aimed at segmenting dissection zones and specifically designed to incorporate different visual prompts, such as scribbles and bounding boxes. Our approach overlays these visual cues directly onto the images, utilizing fine-tuning of the foundational model on a specialized dataset created to handle diverse visual prompt instructions. This shift towards more flexible input methods is intended to significantly improve both the performance of dissection zone segmentation and the overall user experience.

\textbf{Results:} 
We evaluate our approaches using the three experimental setups: in-domain evaluation, evaluation under variability in visual prompts availability, and robustness assessment. By validating our approaches on the ESD-DZSeg dataset, specifically focused on the dissection zone segmentation task of ESD, our experimental results show that our solution outperforms state-of-the-art segmentation methods for this task. To the best of our knowledge, this is the first study to incorporate visual prompt design in dissection zone segmentation. 

\textbf{Conclusion:} 
We introduce the Prompted-based Dissection Zone Segmentation (PDZSeg) model, which is specifically designed for dissection zone segmentation and can effectively utilize various visual prompts, including scribbles and bounding boxes. This model improves segmentation performance and enhances user experience by integrating a specialized dataset with a novel visual referral method that optimizes the architecture and boosts the effectiveness of dissection zone suggestions. Furthermore, we present the ESD-DZSeg dataset for robot-assisted endoscopic submucosal dissection (ESD), which serves as a benchmark for assessing dissection zone suggestions and visual prompt interpretation, thus laying the groundwork for future research in this field. Our code is available at \href{https://github.com/FrankMOWJ/PDZSeg}{https://github.com/FrankMOWJ/PDZSeg}.
}




\maketitle

\section{Introduction}
Endoscopic Submucosal Dissection (ESD) is a surgical procedure employed in the treatment of early-stage gastrointestinal cancers~\cite{chiu2012endoscopic,zhang2020symmetric}. This procedure entails a complex series of dissection maneuvers that require significant skill to determine the dissection zone. In traditional ESD, a transparent cap is employed to retract lesions, which can often obscure the view of the submucosal layer and lead to an incomplete dissection zone. Conversely, our robot-assisted ESD~\cite{gao2024transendoscopic} offers better visualization of the submucosal layer, resulting in a more completed dissection zone by utilizing robotic instruments that enable independent control over retraction and dissection. Achieving successful submucosal dissection requires the careful excision of the lesion or mucosal layer along with the complete submucosal layer while ensuring that both the underlying muscular layer and the mucosal surface remain unharmed. If the electric knife inadvertently contacts tissue outside the designated dissection area, it can lead to damage to the muscle layer, increasing the risk of perforations. Such complications not only elevate the surgical risks but can also complicate the patient’s recovery. Therefore, it is imperative to maintain a precise dissection zone during endoscopic procedures. Effective guidance can help ensure that surgeons are adept at identifying and adhering to appropriate dissection boundaries and enhance the overall safety of endoscopic submucosal dissection (ESD). 


However, the complexity of endoscopic surgical environments presents substantial challenges for the dissection zone segmentation task. Traditional segmentation algorithms often struggle to accurately identify relevant features within these intricate images because unrelated areas can easily mislead them. This tendency for confusion can lead to ambiguous segmentation outcomes, which fail to delineate the dissection area. In addition, during surgery, it is common for surgeons to encounter situations where they must make quick decisions about dissection zones that can be safely operated. Without proper guidance, they may hesitate, leading to indecision that can prolong the procedure and increase the risk of adverse events. In a surgical setting, a seasoned physician with extensive experience and technical proficiency often provides guidance and visual prompts about the area that can be safely dissected to a less experienced colleague. 


Drawing inspiration from the provided visual prompts, we propose training the models with their assistance. While existing deep learning-based segmentation models focus on whole image understanding, a prominent gap exists in achieving region-specific comprehension. Users can supply a visual prompt to indicate the region of interest (ROI), which refers to the specific area within an image that they wish to analyze or isolate. Many visual prompting systems have predominantly relied on regular shapes for visual prompts. However, we recognize the need for a model to interpret a broader spectrum of visual cues. Our research aims to enhance the interaction with models, making it more natural and intuitive. To address these challenges, we present an innovative segmentation model that adapts a foundation model to interpret various visual prompts. This design empowers users to intuitively plot images and engage with the model through natural cues, such as a “bounding box” or “scribble.” By directly overlaying visual prompts onto the RGB image, our approach eliminates the need for intricate prompt encodings while still achieving state-of-the-art performance in the dissection zone segmentation task.


We adopt the foundation models in this work because they have become a groundbreaking technique across multiple domains, utilizing extensive datasets to create generalized representations that can be fine-tuned for various downstream tasks. Their ability to recognize complex patterns and relationships in data makes them a strong foundation for numerous applications, providing an excellent starting point for training models tailored to specific needs. Given the advanced capabilities in extracting unified visual features, we have chosen DINOv2~\cite{oquab2023dinov2} as the foundational backbone for developing the dissection zone segmentation model specifically designed for endoscopic surgery. 


However, one of the challenges with foundation models is their tendency to experience a marked decline in predictive performance when applied to specialized domains. By injecting low-rank matrices into the model's layers, Low-Rank Adaptation (LoRA)~\cite{hu2021lora} allows for quick adjustments without extensive retraining, facilitating the use of foundation models in specialized segmentation tasks. This approach significantly reduces the computational and memory requirements for training, making the deployment of the model more feasible and efficient. 


In summary, our method can streamline the training process by eliminating the need for complex prompt encodings, and achieve more precise delineation of the dissection zone during endoscopic procedures, thereby significantly enhancing the overall effectiveness of the surgical intervention. In this work, Our contributions are summarized as follows:
\begin{itemize}
\item We present a Prompted-based Dissection Zone Segmentation (PDZSeg) model, designed for dissection zone segmentation and tailored to accommodate various visual prompts, including scribbles and bounding boxes. Our methodology involves overlaying these visual prompts directly onto the images, facilitated by fine-tuning the foundation model on our specialized dataset developed to manage a wide range of visual prompt instructions. This transition to more adaptable input methods seeks to enhance the dissection zone segmentation performance and user experience significantly.

\item We propose a novel visual referral method for the dissection zone suggestion task that integrates visual prompts directly onto images, streamlining the model architecture while enhancing performance.

\item We develop the ESD-DZSeg dataset for the dissection zone segmentation task for robot-assisted ESD from ex-vivo porcine models using our custom dual-arm robotic system~\cite{gao2024transendoscopic}. It can be regarded as a benchmark for evaluating dissection zone suggestion and assessing visual prompt interpretation, establishing a foundational framework for future research.

\end{itemize}

\label{sec:intro}

\begin{figure*}[!hbpt]
\centering
\includegraphics[width=1\linewidth]{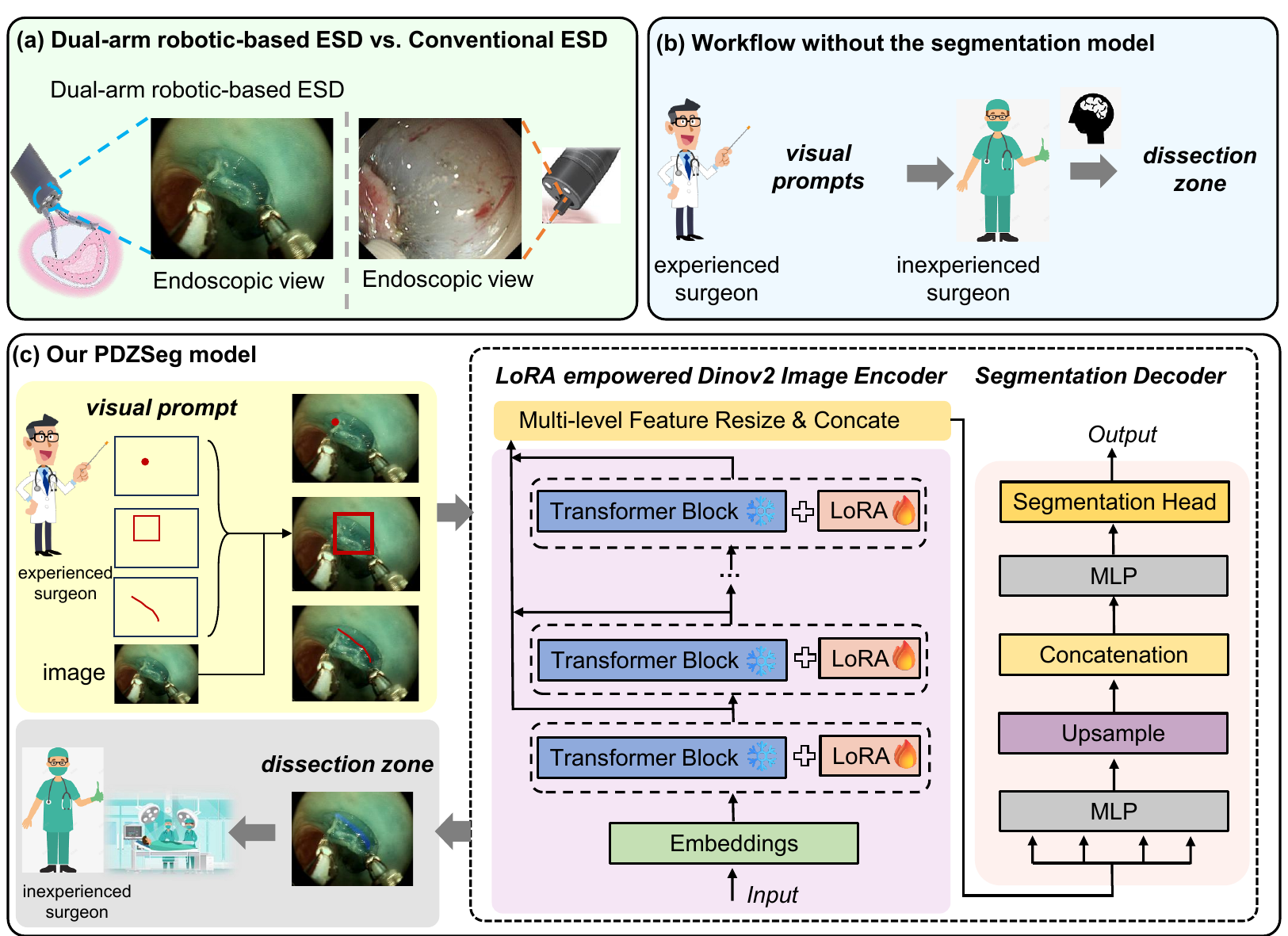}
\caption{Overview. (a) Robotic ESD procedures were recorded from ex-vivo porcine models utilizing our custom dual-arm robotic platform~\cite{gao2024transendoscopic}. Robotic ESD provides improved visualization of the submucosal layer, leading to a more complete dissection zone compared to conventional ESD. (b) The detailed segmentation zone contour guidance is challenging for doctors to provide in real time. (c) We introduce a model, PDZSeg, specifically developed for segmenting dissection zones and capable of integrating different visual prompts provided by the experienced surgeon, such as scribbles and bounding boxes. Our approach incorporates these visual cues directly onto the images, leveraging fine-tuning of the foundational model. Our model then delivers precise dissection zone contours to the inexperienced surgeon.}
\label{fig:overview}
\end{figure*}

\section{Related work}
\textbf{Segmentation models} Traditional segmentation models~\cite{zheng2021rethinking, fan2021rethinking,kirillov2020pointrend,ni2022dnl,chen2017rethinking,zhao2017pyramid} often struggle with complex images. Recent segmentation models have addressed the limitations of traditional methods by incorporating user inputs into the segmentation process. The integration of foundation models~\cite{kirillov2023segment} into segmentation tasks has revolutionized the field. The Segment Anything Model (SAM)~\cite{kirillov2023segment} requires a prompt embedding module to handle visual prompts. The Medical-SAM-Adapter~\cite{wu2023medical} was developed to address SAM's limitations in medical image segmentation, particularly its lack of domain-specific knowledge.




\textbf{Visual prompting systems} Traditional systems have predominantly relied on regular shapes for visual prompts. Many existing approaches~\cite{zhang2023gpt4roi,chen2306shikra} primarily utilize bounding box inputs for visual referrals. While this method is effective in structured environments, it falls short in more dynamic, user-driven interactions where visual prompts may not adhere to regular geometric forms. Recent advancements in interactive segmentation techniques demonstrate the potential for using points or scribbles as input~\cite{kirillov2023segment}. GPT-4V can comprehend various markers~\cite{yang2023dawn}.

\section{Methods}
\subsection{Data collection and annotation}

\begin{figure*}[!hbpt]
\centering
\includegraphics[width=0.7\linewidth]{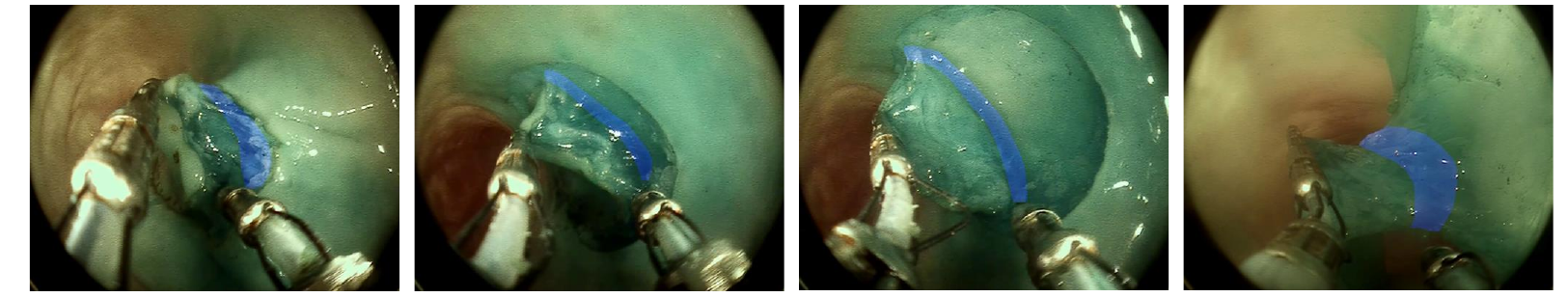}
\caption{Our ESD-DZSeg dataset. The blue areas represent the ground truth of the dissection zone. The complexities of the endoscopic scene understanding task can be observed from the dataset. The features across various regions exhibit significant similarity, and the boundaries between these regions are often ambiguous.}
\label{fig:dataset}
\end{figure*}

In this study, $21$ robotic ESD procedures were collected from ex-vivo porcine models using our custom dual-arm robotic system~\cite{gao2024transendoscopic} in Qilu Hospital. Video recordings were captured at 30 FPS with $1920 \times 1080$ resolution using a flexible dual-channel endoscope. After processing, the endoscopic images were cropped to achieve a final resolution of $1310 \times 1010$. This research specifically targets the submucosal dissection task, which is characterized by extensive interactions with soft tissue and procedural intricacies. Video sequences pertinent to this task were selectively extracted and downsampled to 1 FPS to create the dissection dataset.
We curated a total of $1,849$ images requiring dissection zone recommendations, sourced from $21$ videos. These images were divided into two distinct subsets. Specifically, video 050744, video 030204, video 000032, and video 103115 comprised the test set, which included $369$ images. The remaining videos were allocated to the training set, totaling $1,480$ images.
Expert endoscopists from Qilu Hospital were responsible for annotating the dissection zones for these images utilizing the annotation software LabelBox\footnote{\url{https://labelbox.com/}}. Fig.~\ref{fig:dataset} illustrates our ESD-DZSeg (Dissecon Zone Segmentation for ESD) dataset.

\vspace{-2.0em}
\subsection{Our dissection zone segmentation model}

\subsubsection{Visual prompts design}

Our research aims to enhance the interaction with models, making it more natural and intuitive. Designing visual cues of varying intensities tailored to the needs of less experienced surgeons can significantly improve the confidence and safety of less experienced surgeons during surgery. Thus we provide flexible visual prompt options. Users can provide inputs in various formats, such as:
\textbf{points}: Drawing a single point to indicate a region of interest. \textbf{scribbles (long/short)}: Drawing freehand lines to specify the location of regions of interest (ROIs). \textbf{bounding boxes}: Encapsulating objects or areas within rectangular frames for easier identification. These natural visual prompts serve as cues for our dissection zone segmentation model, directing its attention to the specified ROIs, thus enhancing the efficiency and accuracy of the segmentation process. These visual cues can provided by the more experienced surgeon (see Fig.~\ref{fig:overview}).


\subsubsection{Dissection zone segmentation model}
Essentially, our segmentation model suggests the dissection zone for these inexperienced doctors, emphasizing the detailed delineation of critical areas that may be challenging for experienced professionals to guide in real time. The proposed Prompted-based Dissection Zone Segmentation (PDZSeg) model comprises three primary components: an image encoder, a LoRA efficient training module, and a segmentation decoder. The module processes the RGB raw images attached with the visual prompt and generates the dissection zone masks.

\textbf{Image encoder.} DINOv2~\cite{oquab2023dinov2} is a state-of-the-art self-supervised vision foundation model, designed to learn powerful image representations from large-scale unlabelled datasets. We adopt the pre-trained DINOv2 as our image encoder, which can generate multi-level, rich feature representations that enhance the subsequent segmentation task.
To be specific, an input image $x \in \mathbb{R} ^{H\times W \times C}$ first be embedded into non-overlapping patches $z_k^0 \in \mathbb R ^ {N \times D}$, $1 \leq k \leq N$, where $N = \frac{HW}{p^2}$, $p$ is the size of patch and $D$ is the dimension of each patch, attached with an additional class token $z_0^0 \in \mathbb R ^{1 \times D}$, following the practical in Vision Transformer(ViT)~\cite{dosovitskiy2020vit}. 
The combined tokens $z^0 = [z^0_0, z^0_1, ..., z^0_k]$ are then passed through $M$ transformer blocks to extract intermediate feature representations, denoted as $z^m$ for $1 \leq m \leq M$. At each selected stage, the intermediate feature representations, consisting of both the class token and image tokens, are concatenated along the channel dimension and upsampled by a factor of 4. The features from all the selected stages are then combined to form the final image representation for the decoder.

\textbf{LoRA efficient training module.} For efficient training, we only add the LoRA layer to the $W_q$ and $W_v$ matrices in each of the attention blocks. Specifically, let $x^{t}_{i}$ represent a token embedding at the $t$th transformer block, and $W^{t}_k$, $W^{t}_q$, and $W^{t}_v$ be the key, query, and value projection matrices of the transformer block, respectively. Let $A_q$, $B_q$, $A_v$, and $B_v$ denote the LoRA low-rank matrices corresponding to $W^{t}_q$ and $W^{t}_v$. The attention scores can then be calculated as: $ \text{Att}(Q, K, V) = \text{Softmax} \left( \frac{QK^T}{\sqrt{D}} \right) $, $Q = W_q x + B_q A_q x$, $K = W_k x$, $V  =  W_v x + B_v A_v x$, and $D$ is the numbers of output tokens.



\textbf{Segmentation decoder.} A lightweight All-MLP network is utilized as the segmentation decoder. Each multi-level feature representation is first standardized to the same number of channels by an MLP layer. Then, they are resized back to the resolution of the input image. Afterward, the multi-level feature representations are concatenated and fused by another MLP layer. Finally, an MLP-based segmentation head maps the fused features to a $c$-channel output, where $c$ is the number of classes to be classified.

\section{Experiments}

\subsection{Implementation details}
We implement our models using the PyTorch framework and conduct all experiments on a single NVIDIA RTX 4090 GPU. The images and their corresponding segmentation masks are resized to 532 $\times$ 532. Vit-Base with 12 transformer blocks and feature dimension of 784 is adopted as the backbone of DINOv2~\cite{oquab2023dinov2}. Intermediate outputs from $m = {3, 6, 9, 12}$  transformer blocks are extracted as image representations. A pre-trained checkpoint is loaded for the DINOv2 encoder, while the segmentation decoder is randomly initialized. The encoder is fine-tuned using LoRA~\cite{hu2021lora} with rank of 4, and the decoder is trained from scratch with a learning rate of 0.001. The batch size is set to 8. We train the models for 100 epochs using the Adam~\cite{KingBa15} optimizer with cross-entropy (CE) loss and the Cosine Annealing scheduler. 

\subsection{Results and Analysis}

\begin{table}[]
\centering
\caption{The dissection zone segmentation performance of different models, with and without various visual prompts.}
\scalebox{0.8}{
\begin{tabular}{cccccccc}
\hline
\rowcolor[HTML]{EFEFEF} 
\cellcolor[HTML]{EFEFEF}                                                                           & \cellcolor[HTML]{EFEFEF}                        & \multicolumn{2}{c}{\cellcolor[HTML]{EFEFEF}Dissection zone} & \multicolumn{2}{c}{\cellcolor[HTML]{EFEFEF}No-go zone} & \cellcolor[HTML]{EFEFEF}                                                                      & \cellcolor[HTML]{EFEFEF}                                                                       \\
\rowcolor[HTML]{EFEFEF} 
\multirow{-2}{*}{\cellcolor[HTML]{EFEFEF}\begin{tabular}[c]{@{}c@{}}Visual \\ Prompt\end{tabular}} & \multirow{-2}{*}{\cellcolor[HTML]{EFEFEF}Model} & IoU                          & Dice                         & IoU                        & Dice                      & \multirow{-2}{*}{\cellcolor[HTML]{EFEFEF}\begin{tabular}[c]{@{}c@{}}Mean \\ IoU\end{tabular}} & \multirow{-2}{*}{\cellcolor[HTML]{EFEFEF}\begin{tabular}[c]{@{}c@{}}Mean \\ Dice\end{tabular}} \\ \hline
                                                                                                   & SETR                                            & 43.91                        & 61.02                        & 97.93                      & 98.95                     & 70.92                                                                                         & 79.99                                                                                          \\
                                                                                                   & STDC                                            & 41.92                        & 59.07                        & 97.79                      & 98.88                     & 77.04                                                                                         & 78.97                                                                                          \\
                                                                                                   & Point-Rend                                      & 44.71                        & 61.79                        & 97.65                      & 98.81                     & 71.18                                                                                         & 80.30                                                                                          \\
                                                                                                   & Fast-SCNN                                       & 41.86                        & 59.01                        & 97.77                      & 98.87                     & 69.81                                                                                         & 78.94                                                                                          \\
                                                                                                   & DeepLabv3                                       & 45.95                        & 62.97                        & 97.90                      & 98.94                     & 71.93                                                                                         & 80.95                                                                                          \\
\multirow{-6}{*}{\begin{tabular}[c]{@{}c@{}}Without\\ visual prompt\end{tabular}}                  & Our DZSeg                                       & 47.55                        & 63.01                        & 98.05                      & 99.04                     & 72.80                                                                                         & 81.02                                                                                          \\ \hline
                                                                                                   & SETR                                            & 50.53                        & 67.14                        & 98.18                      & 99.08                     & 74.36                                                                                         & 83.11                                                                                          \\
                                                                                                   & STDC                                            & 51.64                        & 68.11                        & 98.21                      & 99.10                     & 74.93                                                                                         & 83.61                                                                                          \\
                                                                                                   & Point-Rend                                      & 54.49                        & 70.54                        & 98.25                      & 99.12                     & 76.37                                                                                         & 84.83                                                                                          \\
                                                                                                   & Fast-SCNN                                       & 51.51                        & 68.00                        & 98.25                      & 99.12                     & 74.88                                                                                         & 83.56                                                                                          \\
                                                                                                   & DeepLabv3                                       & 55.57                        & 71.44                        & 98.37                      & 99.18                     & 76.97                                                                                         & 85.31                                                                                          \\
                                                                                                   & SAM-Adapter                                     & 41.43                        & 55.44                        & /                          & /                         & \textbf{/}                                                                                    & /                                                                                              \\
\multirow{-7}{*}{\begin{tabular}[c]{@{}c@{}}With point\\ visual prompt\end{tabular}}               & Our PDZSeg                                      & 56.37                        & 71.43                        & 98.46                      & 99.22                     & 77.41                                                                                         & 85.32                                                                                          \\ \hline
                                                                                                   & SETR                                            & 64.48                        & 78.4                         & 98.75                      & 99.37                     & 81.61                                                                                         & 88.89                                                                                          \\
                                                                                                   & STDC                                            & 69.41                        & 81.94                        & 98.90                      & 99.45                     & 84.15                                                                                         & 90.69                                                                                          \\
                                                                                                   & Point-Rend                                      & 67.87                        & 80.86                        & 98.80                      & 99.40                     & 83.33                                                                                         & 90.13                                                                                          \\
                                                                                                   & Fast-SCNN                                       & 66.36                        & 79.78                        & 98.73                      & 99.36                     & 82.55                                                                                         & 89.57                                                                                          \\
                                                                                                   & DeepLabv3                                       & 69.24                        & 81.83                        & 98.90                      & 99.45                     & 84.07                                                                                         & 90.64                                                                                          \\
\multirow{-6}{*}{\begin{tabular}[c]{@{}c@{}}With bounding box\\ visual prompt\end{tabular}}        & Our PDZSeg                                      & 69.85                        & 82.00                        & 98.97                      & 99.48                     & 84.41                                                                                         & 90.73                                                                                          \\ \hline
                                                                                                   & SETR                                            & 62.38                        & 76.83                        & 98.66                      & 99.33                     & 80.52                                                                                         & 88.08                                                                                          \\
                                                                                                   & STDC                                            & 62.65                        & 77.04                        & 98.58                      & 99.28                     & 80.61                                                                                         & 88.16                                                                                          \\
                                                                                                   & Point-Rend                                      & 64.42                        & 78.36                        & 98.74                      & 99.37                     & 81.58                                                                                         & 88.86                                                                                          \\
                                                                                                   & Fast-SCNN                                       & 61.43                        & 76.11                        & 98.60                      & 99.30                     & 80.02                                                                                         & 87.70                                                                                          \\
                                                                                                   & DeepLabv3                                       & 64.15                        & 78.16                        & 98.73                      & 99.36                     & 81.44                                                                                         & 88.76                                                                                          \\
\multirow{-6}{*}{\begin{tabular}[c]{@{}c@{}}With short scribble\\ visual prompt\end{tabular}}      & Our PDZSeg                                      & 65.45                        & 78.41                        & 98.83                      & 99.41                     & 82.14                                                                                         & 88.91                                                                                          \\ \hline
                                                                                                   & SETR                                            & 70.85                        & 82.94                        & 98.97                      & 99.48                     & 84.91                                                                                         & 91.21                                                                                          \\
                                                                                                   & STDC                                            & 70.19                        & 82.49                        & 98.94                      & 99.47                     & 84.56                                                                                         & 90.98                                                                                          \\
                                                                                                   & Point-Rend                                      & 74.73                        & 85.54                        & 99.17                      & 99.58                     & 86.95                                                                                         & 92.56                                                                                          \\
                                                                                                   & Fast-SCNN                                       & 71.87                        & 83.63                        & 99.04                      & 99.52                     & 85.46                                                                                         & 91.58                                                                                          \\
                                                                                                   & DeepLabv3                                       & 72.30                        & 83.93                        & 99.08                      & 99.54                     & 85.69                                                                                         & 91.73                                                                                          \\
\multirow{-6}{*}{\begin{tabular}[c]{@{}c@{}}With long scribble \\ visual prompt\end{tabular}}      & Our PDZSeg                                            & 74.06                        & 84.30                        & 99.15                      & 99.56                     & 86.60                                                                                         & 91.93                                                                                          \\ \hline
\end{tabular}
}
\label{main_table}
\end{table}

\textbf{Model evaluation with and without visual prompts}
To evaluate the performance of our proposed method, several State-of-the-art(SOTA) models, such as SETR~\cite{zheng2021rethinking}, STDC~\cite{fan2021rethinking}, Point-Rend~\cite{kirillov2020pointrend}, Fast-SCNN~\cite{poudel2019fast}, DeepLabv3~\cite{chen2017rethinking}, and Medical-SAM-Adapter~\cite{wu2023medical} are used as our baselines. The area outside the dissection zone is termed the “No-go zone”.


Table~\ref{main_table} summarizes the segmentation performance of various models, highlighting the following key findings: (a) When we revise our model (PDZSeg) to a version that does not utilize visual cues (DZSeg), it still achieves better segmentation results than most other models. (b) These baseline models do not inherently incorporate visual prompt designs; however, their performance has significantly improved when combined with our visual prompt approach. (c) The results obtained with visual prompts are substantially better than those without. This indicates that segmentation performance improves notably when visual prompts direct the model towards specific regions of interest. (d) Among the four forms of visual prompts we designed, the model employing long scribble prompts exhibits the highest performance, followed by the model using bounding box prompts and short scribble prompts. The model utilizing point prompts produces the least favorable results.

\begin{figure*}[!hbpt]
\centering
\includegraphics[width=1\linewidth]{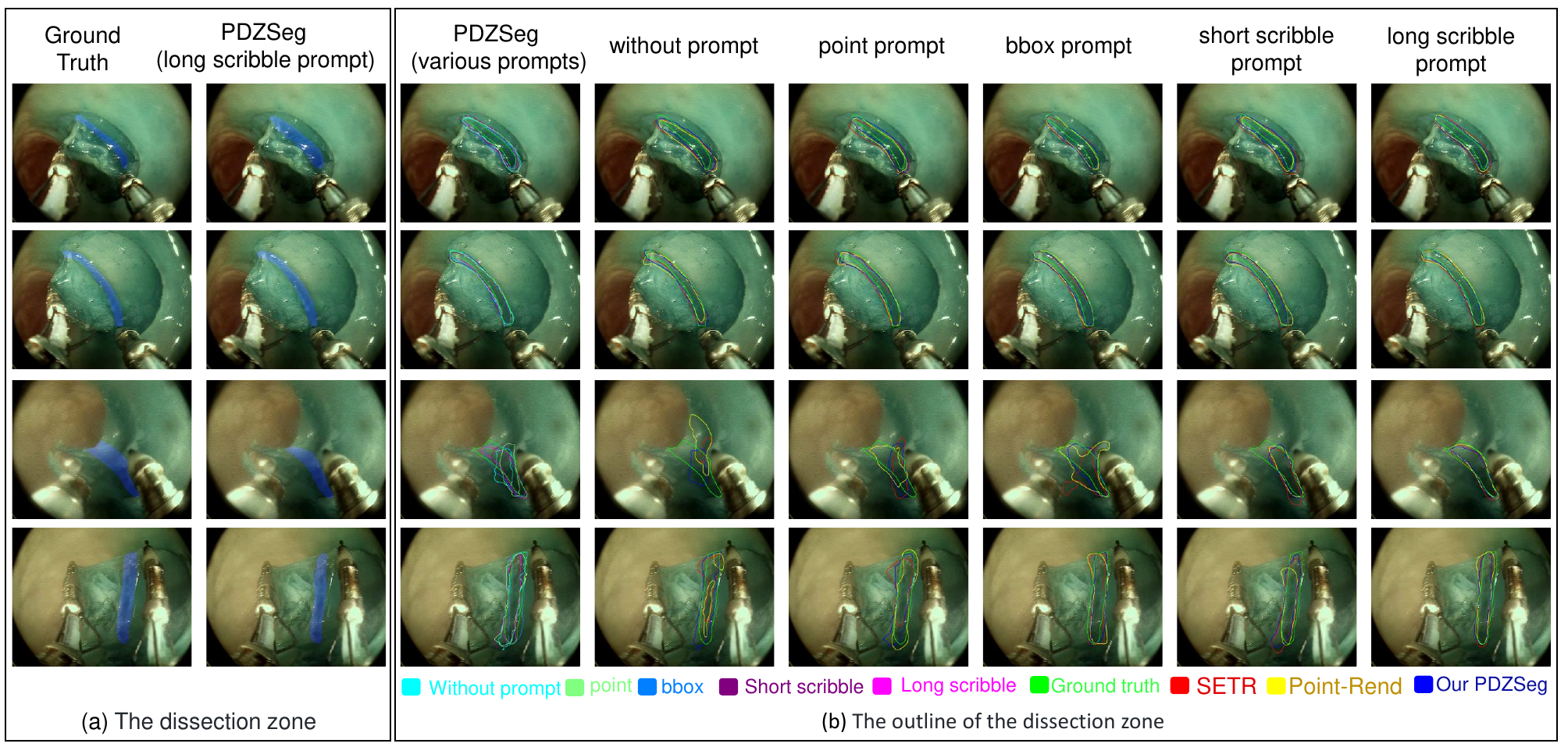}
\caption{Results visualization. The first two columns display the ground truth and the segmentation masks predicted by our model. To enhance the comparison, we have extracted the outlines of the segmentation masks, which more clearly highlight the boundaries of the dissection zones. The differently colored contours represent the results obtained from corresponding methods.}
\label{fig:results}
\end{figure*}

\textbf{Model evaluation under variability in visual prompt availability:}
We also integrate a diverse dataset that includes both prompted and non-prompted scenarios to represent the complexities inherent in real-world clinical environments more accurately. The integration ratios of prompt and non-prompt data are 6:4, 5:5, and 4:6, respectively. In clinical practice, even seasoned physicians may encounter situations where they cannot provide comprehensive prompts for every case, leading to variability in the information available for decision-making. The assessment outcomes are displayed in Table~\ref{ab1}. Moreover, we also integrate a blend of data that includes a variety of prompt types alongside non-prompted scenarios, reflecting the range of responses that physicians might offer in different contexts. Data prompts with long scribbles, short scribbles, bounding boxes, and data with no prompts account for $25\%$. This variability in prompt types is crucial, as it acknowledges that clinicians may have differing levels of insight or detail depending on the specific circumstances of each case. The evaluation findings are illustrated in Table~\ref{ab2}.



\begin{table}[]
\caption{Our model evaluation under a diverse dataset of prompted and non-prompted scenarios in ratios of 6:4, 5:5, and 4:6.}
\centering
\scalebox{0.8}{
\begin{tabular}{cccccccc}
\hline
\rowcolor[HTML]{EFEFEF} 
\cellcolor[HTML]{EFEFEF}                                                                           & \cellcolor[HTML]{EFEFEF}                                 & \multicolumn{2}{c}{\cellcolor[HTML]{EFEFEF}Dissection zone} & \multicolumn{2}{c}{\cellcolor[HTML]{EFEFEF}No-go zone} & \cellcolor[HTML]{EFEFEF}                                                                      & \cellcolor[HTML]{EFEFEF}                                                                       \\
\rowcolor[HTML]{EFEFEF} 
\multirow{-2}{*}{\cellcolor[HTML]{EFEFEF}\begin{tabular}[c]{@{}c@{}}Visual \\ Prompt\end{tabular}} & \multirow{-2}{*}{\cellcolor[HTML]{EFEFEF}with : without} & IoU                          & Dice                         & IoU                        & Dice                      & \multirow{-2}{*}{\cellcolor[HTML]{EFEFEF}\begin{tabular}[c]{@{}c@{}}Mean \\ IoU\end{tabular}} & \multirow{-2}{*}{\cellcolor[HTML]{EFEFEF}\begin{tabular}[c]{@{}c@{}}Mean \\ Dice\end{tabular}} \\ \hline
                                                                                                   & 6--4                                                     & 73.41                        & 83.98                        & 99.12                      & 99.56                     & 86.27                                                                                         & 91.76                                                                                          \\
                                                                                                   & 5--5                                                     & 73.46                        & 83.87                        & 99.13                      & 99.56                     & 86.31                                                                                         & 91.72                                                                                          \\
\multirow{-3}{*}{\begin{tabular}[c]{@{}c@{}}long scribble\\ visual prompt\end{tabular}}            & 4--6                                                     & 71.81                        & 83.08                        & 99.07                      & 99.53                     & 85.45                                                                                         & 91.3                                                                                           \\ \hline
                                                                                                   & 6--4                                                     & 58.04                        & 71.49                        & 98.52                      & 99.25                     & 78.28                                                                                         & 85.37                                                                                          \\
                                                                                                   & 5--5                                                     & 54.72                        & 68.7                         & 98.43                      & 99.2                      & 76.58                                                                                         & 83.97                                                                                          \\
\multirow{-3}{*}{\begin{tabular}[c]{@{}c@{}}short scribble\\ visual prompt\end{tabular}}           & 4--6                                                     & 52.51                        & 66.91                        & 98.32                      & 99.15                     & 75.41                                                                                         & 83.03                                                                                          \\ \hline
                                                                                                   & 6--4                                                     & 58.96                        & 72.04                        & 98.89                      & 99.43                     & 78.92                                                                                         & 85.73                                                                                          \\
                                                                                                   & 5--5                                                     & 56.85                        & 70.87                        & 98.45                      & 99.21                     & 77.65                                                                                         & 85.04                                                                                          \\
\multirow{-3}{*}{\begin{tabular}[c]{@{}c@{}}bounding box\\ visual prompt\end{tabular}}             & 4--6                                                     & 55.2                         & 68.93                        & 98.39                      & 99.18                     & 76.79                                                                                         & 84.06                                                                                          \\ \hline
\end{tabular}
}
\label{ab1}
\end{table}

\begin{table}[]
\caption{Our model evaluation under a dataset that includes various prompt types and non-prompted scenarios, with each type comprising 25\% of the data.}
\centering
\scalebox{0.8}{
\begin{tabular}{cccccccc}
\hline
\rowcolor[HTML]{EFEFEF} 
\cellcolor[HTML]{EFEFEF}                                                                                        & \cellcolor[HTML]{EFEFEF}                                                                                      & \multicolumn{2}{c}{\cellcolor[HTML]{EFEFEF}Dissection zone} & \multicolumn{2}{c}{\cellcolor[HTML]{EFEFEF}No-go zone} & \cellcolor[HTML]{EFEFEF}                                                                      & \cellcolor[HTML]{EFEFEF}                                                                       \\
\rowcolor[HTML]{EFEFEF} 
\multirow{-2}{*}{\cellcolor[HTML]{EFEFEF}\begin{tabular}[c]{@{}c@{}}Visual prompt \\ in train set\end{tabular}} & \multirow{-2}{*}{\cellcolor[HTML]{EFEFEF}\begin{tabular}[c]{@{}c@{}}Visual prompt \\ in val set\end{tabular}} & IoU                          & Dice                         & IoU                        & Dice                      & \multirow{-2}{*}{\cellcolor[HTML]{EFEFEF}\begin{tabular}[c]{@{}c@{}}Mean \\ IoU\end{tabular}} & \multirow{-2}{*}{\cellcolor[HTML]{EFEFEF}\begin{tabular}[c]{@{}c@{}}Mean \\ Dice\end{tabular}} \\ \hline
                                                                                                                & long scribble                                                                                                     & 70.68                        & 81.86                        & 99.00                         & 99.49                     & 84.84                                                                                         & 90.67                                                                                          \\
                                                                                                                & short scribble                                                                                                     & 60.18                        & 74.84                        & 98.65                      & 99.31                     & 79.41                                                                                         & 87.08                                                                                          \\
                                                                                                                & bounding box                                                                                                          & 64.92                        & 78.27                        & 98.78                      & 99.38                     & 81.85                                                                                         & 88.82                                                                                          \\
\multirow{-4}{*}{\begin{tabular}[c]{@{}c@{}}mix visual\\ prompt\end{tabular}}                                   & without prompt                                                                                                & 44.23                        & 59.13                        & 97.96                      & 98.96                     & 71.09                                                                                         & 79.05                                                                                          \\ \hline
\end{tabular}
}
\label{ab2}
\end{table}

\begin{figure*}[!hbpt]
\centering
\includegraphics[width=0.7\linewidth]{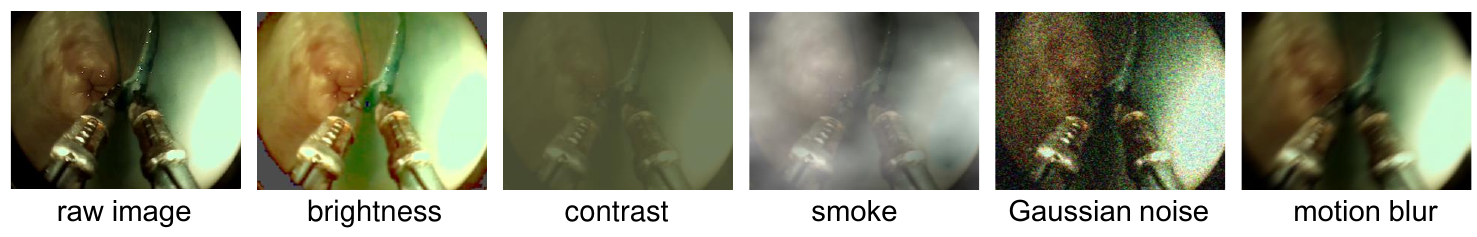}
\caption{Corrupted images at a severity level of 3.}
\label{fig:robustness}
\end{figure*}

\textbf{Robustness evaluation} As illustrated in Fig.~\ref{fig:robustness}, we apply five types of image corruption techniques to our validation set, each at a severity level of 3. These techniques include Gaussian noise, motion blur, smoke, brightness, and contrast adjustments~\cite{hendrycks2019benchmarking}. The evaluation results are presented in Table~\ref{label_robustness}.

\begin{table}[]
\centering
\caption{Performance of dissection zone segmentation under five different image corruption techniques.}
\scalebox{0.7}{
\begin{tabular}{cccccccccccc}
\hline
\rowcolor[HTML]{EFEFEF} 
\cellcolor[HTML]{EFEFEF}                                                                          & \cellcolor[HTML]{EFEFEF}                        & \multicolumn{2}{c}{\cellcolor[HTML]{EFEFEF}Gaussian noise} & \multicolumn{2}{c}{\cellcolor[HTML]{EFEFEF}Motion blur} & \multicolumn{2}{c}{\cellcolor[HTML]{EFEFEF}Smoke} & \multicolumn{2}{c}{\cellcolor[HTML]{EFEFEF}Contrast} & \multicolumn{2}{c}{\cellcolor[HTML]{EFEFEF}Brightness} \\
\rowcolor[HTML]{EFEFEF} 
\multirow{-2}{*}{\cellcolor[HTML]{EFEFEF}\begin{tabular}[c]{@{}c@{}}Visual\\ Prompt\end{tabular}} & \multirow{-2}{*}{\cellcolor[HTML]{EFEFEF}Model} & IoU                          & Dice                        & IoU                        & Dice                       & IoU                     & Dice                    & IoU                       & Dice                     & IoU                        & Dice                      \\ \hline
                                                                                                  & Point-Rend                                      & 1.94                         & 3.81                        & 22.28                      & 36.44                      & 5.58                    & 10.56                   & 0.51                      & 1.02                     & 37.08                      & 54.10                     \\
\multirow{-2}{*}{\begin{tabular}[c]{@{}c@{}}without\\ visual\\ prompt\end{tabular}}               & Our PDZSeg                                      & 21.45                        & 31.62                       & 25.15                      & 36.01                      & 24.22                   & 36.07                   & 37.30                     & 50.44                    & 46.21                      & 60.61                     \\ \hline
                                                                                                  & Point-Rend                                      & 13.76                        & 24.19                       & 37.60                      & 54.65                      & 27.19                   & 42.75                   & 23.44                     & 37.97                    & 49.41                      & 66.14                     \\
\multirow{-2}{*}{point}                                                                           & Our PDZSeg                                      & 24.15                        & 36.41                       & 35.34                      & 50.11                      & 28.58                   & 44.57                   & 44.68                     & 61.24                    & 54.21                      & 68.81                     \\ \hline
                                                                                                  & Point-Rend                                      & 16.21                        & 27.90                       & 55.65                      & 71.51                      & 45.00                   & 62.07                   & 43.76                     & 60.88                    & 56.56                      & 74.66                     \\
\multirow{-2}{*}{\begin{tabular}[c]{@{}c@{}}bounding \\ box\end{tabular}}                         & Our PDZSeg                                      & 43.35                        & 56.40                       & 57.14                      & 72.09                      & 60.06                   & 75.01                   & 65.54                     & 78.62                    & 68.70                      & 81.07                     \\ \hline
                                                                                                  & Point-Rend                                      & 4.98                         & 9.49                        & 45.27                      & 62.32                      & 42.58                   & 59.73                   & 44.03                     & 61.14                    & 56.13                      & 71.90                     \\
\multirow{-2}{*}{\begin{tabular}[c]{@{}c@{}}short \\ scribble\end{tabular}}                       & Our PDZSeg                                      & 45.34                        & 57.43                       & 58.81                      & 72.73                      & 59.67                   & 73.98                   & 61.74                     & 75.85                    & 64.66                      & 77.53                     \\ \hline
                                                                                                  & Point-Rend                                      & 2.89                         & 5.61                        & 65.93                      & 79.47                      & 62.28                   & 76.76                   & 64.90                     & 78.72                    & 70.66                      & 82.81                     \\
\multirow{-2}{*}{\begin{tabular}[c]{@{}c@{}}long \\ scribble\end{tabular}}                        & Our PDZSeg                                      & 64.91                        & 77.11                       & 70.09                      & 80.80                      & 70.73                   & 81.21                   & 72.01                     & 82.57                    & 73.52                      & 83.41                     \\ \hline
\end{tabular}
\label{label_robustness}
}
\end{table}

\section{Conclusion}
We introduce the Prompted-based Dissection Zone Segmentation (PDZSeg) model, which is specifically designed for dissection zone segmentation and accommodates various visual prompts such as scribbles and bounding boxes. By overlaying these visual prompts onto images and fine-tuning the foundation model on a specialized dataset, we aim to significantly enhance both segmentation performance and user experience. Our novel visual referral method further streamlines the model architecture while improving effectiveness in dissection zone suggestion tasks. Additionally, we have developed the ESD-DZSeg dataset for robot-assisted endoscopic submucosal dissection (ESD) using ex-vivo porcine models with our custom dual-arm robotic system. This dataset serves as a benchmark for evaluating dissection zone suggestions and visual prompt interpretation, establishing a foundational framework for future research in this domain.

\backmatter

\bmhead{Acknowledgments}
This work was supported in part by the Hong Kong Research Grants Council (RGC) Collaborative Research Fund under Grant CRF C4026-21GF; in part by the STIC Shenzhen-Hong Kong-Macau Technology Research Programme (Type C) under Grant 202108233000303; and in part by the Key Project 2021B1515120035 (B.02.21.00101) of the Regional Joint Fund Project of the Basic and Applied Research Fund of Guangdong Province.

\bibliography{sn-bibliography}

\end{document}